# Simplifying the Bible and Wikipedia Using Statistical Machine Translation
## (Toward a Text Synthesizer via Machine Translation)

Yohan Jo
Language Technologies Institute
yohanj@cs.cmu.edu

Summary

I started this work with the hope of generating a text synthesizer (like a musical synthesizer) that can imitate certain linguistic styles. Most of the report focuses on text simplification using statistical machine translation (SMT) techniques. I applied MOSES to a parallel corpus of the Bible (King James Version and Easy-to-Read Version) and that of Wikipedia articles (normal and simplified). I report the importance of the three main components of SMT—phrase translation, language model, and recording—by changing their weights and comparing the resulting quality of simplified text in terms of METEOR and BLEU. Toward the end of the report will be presented some examples of text "synthesized" into the King James style.

1. INTRODUCTION

Text simplification has been investigated with the aim to help language learners (Petersen & Ostendorf, 2007), the deaf (Inui et al., 2003), aphasic patients (Devlin & Tait, 1998), people in different reading levels (De Belder & Moens, 2010), etc. Approaches to text simplification can be grouped into four categories: lexical simplification, syntactic simplification, explanation generation, and statistical machine translation (Shardlow, 2014). Machine translation (MT) approaches view text simplification as monolingual text-to-text generation. A big advantage of these approaches is that they already employ various textual features to convert one type of text to another. For example, they learn the rules of lexical, phrasal, and syntactic changes, and reordering. They may produce less fluent and adequate text than rule-based approaches do, but MT approaches attempt to compensate the weakness using several techniques such as language models. MT-based text simplification has two major challenges: data and evaluation. MT basically requires a parallel corpus, i.e., a pair of normal text and simple text in this case. It is hard, however, to obtain a large amount of normal-simple text pairs. Evaluation is also a big issue. Conventional metrics for MT that measure similarity in meaning and fluency are not optimal for text simplification, which values readability. In this work, I focus on the first challenge, i.e., data limitation, by introducing a new resource: the Bible. Most previous work depends on Wikipedia with its normal version and simplified version has been used extensively for text simplification. It is worthwhile to explore new types of parallel corpora to better understand the nature of text simplification.

The Bible has many translations with different levels of reading difficulty. Some versions are written in an archaic writing style (e.g., King James Version), some use ordinary modern language (e.g., New International Version), and some are meant to be easy to read (e.g., Easy-to-

Read Version). The main task in this work is text simplification using the King James Version and Easy-to-Read version. These two versions lead to not only text simplification but also a stylistic change from archaic language to modern language. In this paper, I will also show some translation results by switching the source and target languages, i.e., from modern to archaic. This process may be called text complexification, but my intent is rather to see the possibility of a general "text synthesizer" that makes certain stylistic changes to plain text. The diverse translations of the Bible open a way to for this.

This work is focused on qualitative analysis of text simplification, rather than introducing new MT techniques. I use Moses' default phrase-based model and tree-to-tree model. The evaluation based on METEOR and BLEU show that different components in the models have different importance. The language model component rather harmed results, probably because of the small size of data. The phrase translation and reordering components show consistent improvement, but the different was not as noticeable as the damage by the language model component. The phrase-based model consistently outperforms the syntax-based model. Throughout the paper, I will make a lot of comparison between the Bible and Wikipedia to emphasize that different corpora can lead to different characteristics of text simplification.

2. RELATED WORK

This section introduces previous work on text simplification based on statistical machine translation (SMT). We focus especially on three aspects: data, model, and evaluation. Please refer to the survey paper (Shardlow, 2014) for a more comprehensive review of this field.

One of the main obstacles for STM is data, because SMT requires a parallel corpus for training, i.e., a pair of normal text and simple text. There are few sources of parallel corpora. Most work in this domain leverages Wikipedia and Simple English Wikipedia[1] (Kauchak, 2013; Wubben et al., 2012; Zhu et al., 2010; Coaster & Kauchak, 2011a; Coaster & Kauchak, 2011b). Another paper uses documents from German websites (Klaper et al., 2013), but the number of documents is only about 250, which is too small to train a MT model. Therefore, the potential of new data is worth investigating. In this work, I introduce the Bible as a new resource for text simplification. To the best of my knowledge, there is no published work using the Bible for text simplification. By using the Bible, we can obtain new and interesting insights that are not available from Wikipedia. We will see that Wikipedia is only one of many resources that have different characteristics.

Phrase-based MT models are the most widely used for text simplification. Additional functionalities such as allowing deletion from Moses' phrase-based model (Coaster & Kauchak, 2011a) or reranking candidates to pick more different translation from the source text (Wubben et al., 2012) showed effectiveness in Wikipedia simplification. A tree-based model that covers splitting, dropping, reordering, and substitution (Zhu et al., 2010) has also been examined. In some work, readability features—e.g., sentence length and number of long words—were incorporated into the phrase-based model (Stymne et al., 2013). Since it is not the main focus of this work to develop a new technique, I use Moses' default models, one phrase-based and one syntax-based.

---

[1] http://www.wikipedia.org and http://simple.wikipedia.org

Evaluation is another challenge in text simplification in general. There is no standard way to evaluate the quality of simplified text. A lot of previous studies conducted evaluation with specified target readers (Devlin and Tait, 1998; Kandula et al., 2010; Napoles et al., 2011; Inui, 2003). One type of automatic evaluation is psychology-based measures, e.g., average sentence length, proportion of proper names (Stymne et al., 2013). However, these metrics hardly evaluate fluency and are too naïve. Another type of metrics use reference text. Some studies introduce rather simple metrics such as F1 score and edit distance (Coaster & Kauchak, 2011a), and some use standard MT evaluation methods such as BLEU (Coster and Kauchak, 2011a). In this work, I use two standard MT evaluation methods: METEOR (Denkowski and Lavie, 2014) and BLEU (Papineni et al., 2002). MT evaluation methods may be too sensitive to the reference text and do not accept other possible translation. There has been little effort to examine MT evaluation methods for text simplification.

## 3. METHOD

This section describes two machine translation (MT) models I use: phrase-based model and syntax-based model. These models are trained with normal text (source language) and simple text (target language). The two parallel corpora used, i.e., the Bible and Wikipedia, are also described in this section.

### 3.1. MT Models

The following figure shows the overview of phrase-based model and syntax-based model (i.e., tree-to-tree model).

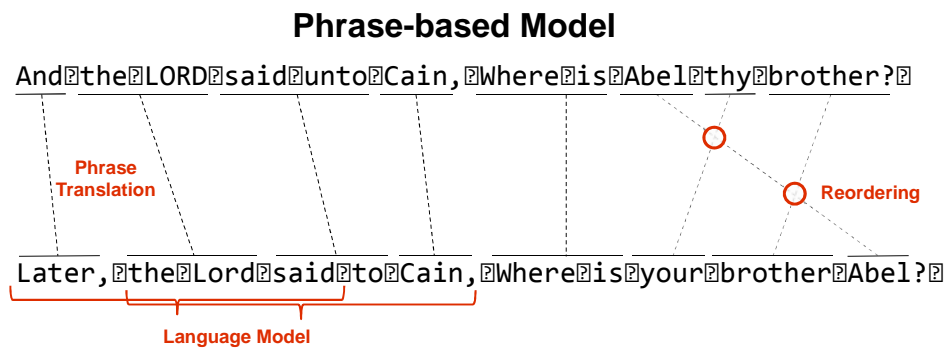

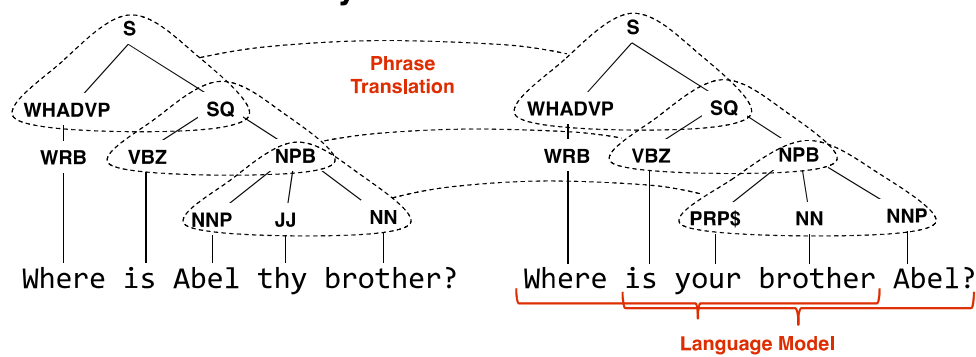

Syntax-based Model

These models have basically three components: phrase translation, language model, and reordering. Although Moses provides more options, e.g., word length penalty, only these three components are used in this work. Phrase translation is translation between phrases. This includes conversion from an archaic word (e.g., "thy") to a modern word (e.g., "your") and word drops within a phrase. Syntactic rules are seen as phrase translation in syntax-based models (e.g., from "NPB → NNP JJ NN" to "NPB → PRP$ NN NNP"). Language models takes into account the probability of n-grams (e.g., the probability of "the Lord said to Cain"). Reordering penalizes or facilitates the change of phrase orders. Syntax-based models do not consider reordering. Overall, the score of a translation is calculated as:

$$\text{score} = w_1 f_{\text{phrase}}(f|e) + w_2 f_{\text{lm}}(e) + w_3 f_{\text{reorder}}(e, f)$$.

One of the main goals in this work is to see the importance of each component. We can verify this by changing the component weights and compare the quality of the results. Likewise, the usefulness of each model will be compared.

3.2. Data

Two datasets are used for text simplification: the Bible and Wikipedia. The basic statistics are summarized in the following table.

|  | Bible | Wikipedia |
|---|---|---|
| Normal Text | King James Version[2]<br><br>• Authorized by King James I in 1604<br>• Most widely read translation in the U.S. (Goff et al., 2014) | www.wikipedia.org<br><br>• Free-access, free content Internet encyclopedia |
| Simple Text | Easy-to-Read Version[3]<br><br>• Originally published as the English Version for the Deaf | simple.wikipedia.org<br><br>• Wikipedia encyclopedia with simpler sentences & grammar |

---

[2] https://www.biblegateway.com/versions/King-James-Version-KJV-Bible/

[3] https://www.biblegateway.com/versions/Easy-to-Read-Version-ERV-Bible/

|   |   |   |
|---|---|---|
|   | (EVD) by BakerBooks<br>• Revised by World Bible Translation Center in 2004<br>• Uses broader vocabulary |   |
| Example | Example 1<br>• Know ye that the LORD he is God: it is he that hath made us.<br>• Know that the Lord is God. He made us.<br>Example 2<br>• So shall my righteousness answer for me in time to come.<br>• In the future, you can easily see if I am honest. | Example 1<br>• Modern decks vary in size, but most are 7 to 10.5 inches wide.<br>• Modern decks are different in size. Most are 7 to 10.5 inches wide.<br>Example 2<br>• Hermes was born on Mount Cyllene in Arcadia.<br>• He was born on Mount Cyllene. |
| Number of aligned verses/sentences | • Before preprocessing: 30,346<br>• After preprocessing: 29,883 | • Before preprocessing: 167,689<br>• After preprocessing: 165,609 |
| Average number of words per verse/sentence (after preprocessing) | • Normal: 28.87 (SD: 11.85)<br>• Simple: 30.66 (SD: 13.51) | • Normal: 25.04 (SD: 12.42)<br>• Simple: 22.69 (SD: 11.11) |
| Percentage of "normal = simple" sentences | • Before preprocessing: 0.1%<br>• After preprocessing: 0.1% | • Before preprocessing: 31.5%<br>• After preprocessing: 29.8% |
| Characteristics | Severe changes in words and syntax | Mostly word drop or no change |
| Hypothesis | • Syntax-based model would work better<br>• Reordering would be important | • Phrase-based model would work better<br>• Reordering would be harmful |

### 3.2.1. Bible

To the best of my knowledge, there is no published work that uses the Bible for English text simplification. This work leverages the King James Version (KJV) as normal text and the Easy-to-Read Version (ERV) as simple text. The KJV has not only a more difficult vocabulary, but also unique stylistic characters. It uses a lot of archaic words (e.g., "thou", "thee", "thy", "ye", "hast", "dwelleth") and prose rhythm. The ERV, in contrast, has simpler and more straightforward sentences. Note that the ERV has a greater number of words in verses on average than the KJV does. This is because the ERV splits a sentence into shorter sentences while keeping its original meaning, that is, it is wordier than the KJV. On the contrary, the Simple Wikipedia has fewer words on average, because it drops less important words, losing some original meaning. The Bible cannot do that.

Since syntactic structures and vocabulary are considerably different between the KJV and ERV, I hypothesized that syntax-based models would work better than phrase-based models. It was also expected that reordering would play an important role in this corpus.

### 3.2.2. Wikipedia

Wikipedia and its simplified version have been widely used in text simplification. The Simplified English Wikipedia (SEW) uses simpler sentences and grammar, and some articles are written only in Basic English. However, the change is not significant. As the table shows, 31.5% of the original sentences remain the same in the SWE, and structural change is not common either. Most of the changes in the SEW are word drops. This is very different from the Bible case, because the SEW is not strict in keeping the original meaning. Simplified text has two fewer words on average.

Since most changes are local and related to word drops in Wikipedia, I hypothesized that phrase-based models would work better than syntax-based models, and reordering might be harmful.

## 4. EXPERIMENTAL SETTINGS

### 4.1. Training MT Models

This work uses the implementation of phrase-based model and syntax-based (tree-to-tree) model provided by Moses (Koehn, 2007), an open-source statistical machine translation toolkit. Moses also provides scripts for preprocessing, e.g., tokenization, case conversion (truecaser), language model learning, etc. The following figures shows the training process and related script names.

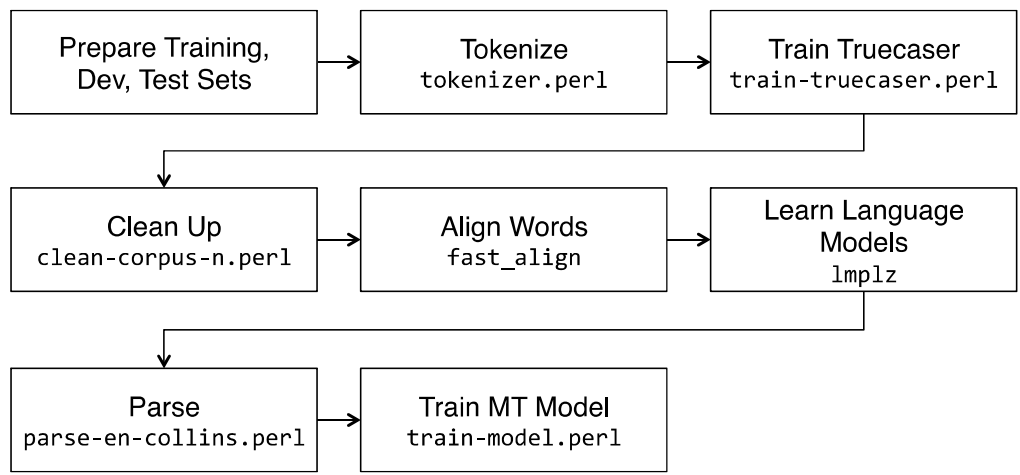

(1) Prepare Training, Dev, Test Sets

This process splits the corpus into three sets: training (80%), dev (10%), and test (10%). The dev set is for tuning parameters.

(2) Tokenize

This process tokenizes words and punctuation marks. This has to be done for all the training, dev, and test sets, because the trained model does not perform tokenization by itself.

(3) Train Truecaser

This process adjusts the capitalization of each word in the data. The first word in a sentence is usually capitalized even if it is not a special word. Since the same word can be treated different depending on capitalization, words should be normalized. Truecaser in Moses automatically lowercases capitalized words when it seems appropriate. This process is done only on the training data, since the trained model does perform this on input text.

(4) Clean Up

This process removes too short or long sentences and adjusts space between words. I retained sentences whose length is between 1 and 70.

(5) Align Words

This process makes word alignment between normal text and simple text. Moses includes GIZA++, but I instead used Fast Align (Dyer et al., 2013), which is faster and easier to use. I used one direction alignment using "-d -o -v" option.

(6) Learn Language Models

This process learns language models. In machine translation, language models are used to make output text fluent. However, in text simplification, it is not enough for output text to be fluent, but the output should also be simplified text. That is, the language model used should represent simplified text styles. Therefore, I did not use external language models (e.g., Europarl), but I rather trained language models up to 5-grams on the training simple text.

(7) Parse

This process identifies the part-of-speech tag of each token and makes parse trees. This is needed only for syntax-based models. The training (normal & simple), dev (normal), and test (normal) sets should be parsed. I used COLLINS-PARSER[4] and MXPOST[5].

(8) Train MT Model

---

[4] http://www.cs.columbia.edu/~mcollins/code.html

[5] http://www.statmt.org/moses/?n=Moses.ExternalTools#ntoc9

This process trains phrase tables for phrase-based models or syntactic rules for syntax-based models. Since I make word alignment using Fast Align, I can train a model from step 4, i.e., getting the lexical translation tables. For syntax-based models, I used the options "--hierarchical --glue-grammar --source-syntax --target-syntax".

4.2. Data Processing

4.2.1. Bible

I use the text of the KJV compiled by Project Gutenberg[6]. I removed text other than verses (e.g., preface). For the ERV, I crawled the website http://bibleabc.net/bible/erv/. Most chapter and verse numbers are marked with specific HTML tags, but there are exceptions, too. For example, some verses are under the same label (e.g., "11-14"), and some chapter numbers are missing. In these cases, I tried my best to correct information, but there are some verses I could not recover automatically. Eventually, I managed to align 30,346 verses out of the entire 31,102 verses in the Bible. (I used verse alignment instead of sentence alignment.) The ERV requires some preprocessing. First, double quotes were removed, since the KJV does not have double quotes. Second, all contents in parentheses were removed; the ERV uses parentheses to add explanations. Third, asterisks were removed, which mark footnotes. After cleaned up by Moses' clean-corpus-n.perl script, total 29,883 verses remained.

4.2.2. Wikipedia

I used a publicly available parallel corpus for Wikipedia (Kauchak, 2013). This dataset has 167,689 aligned sentence pairs. After cleaned up by Moses' clean-corpus-n.perl script, total 165,609 sentences remained.

4.3. Evaluation

I adopt two automatic evaluation methods: METEOR (Denkowski and Lavie, 2014) and BLEU (Papineni et al., 2002). METEOR considers four components: exact word match, stem match, synonym match, and paraphrase. I excluded synonym match, because including synonyms conflicts with the aim of simplification. That is, to change into proper synonyms is the responsibility of text simplification.

5. RESULTS

5.1. Example Output

In this section, I randomly choose and show sentences from the results on the testing set.

5.1.1. Bible

---

[6] http://www.gutenberg.org/

| KJV (Normal) | ERV (Simple) | Phrase-based | Syntax-based |
| --- | --- | --- | --- |
| This is an evil among all things that are done under the sun , that there is one event unto all : yea , also the heart of the sons of men is full of evil , and madness is in their heart while they live , and after that they go to the dead . | Of all the things that happen in this life , the worst thing is that all people end the same way . But it is also very bad that people always think evil and foolish thoughts . And those thoughts lead to death . | this is an evil among all the things that are done in the sun , that there is one event to all people . Yes , even the heart of the sons of men is full of evil , and it is crazy in their hearts as long as they live . and after that they go to the dead . | this that are not true . bad things among all the things that are done under the sun . that there is only event to all with . then God , think for the sons that people is full of evil and crazy in their hearts while they live , and chased after people that were going to destruction . |
| So will I make my fury toward thee to rest , and my jealousy shall depart from thee , and I will be quiet , and will be no more angry . | Then I will stop being angry and jealous . I will calm down . I won 't be angry any more . | , I will make my anger for you to rest , and my strong feelings will stay away from you , and I will be quiet , there will be no more angry . | , I make my people punished for you . to peace . people , and my people that will run away from you . I will be quiet and will be more angry . |
| I thank my God always on your behalf , for the grace of God which is given you by Jesus Christ ; | I always thank my God for you because of the grace that God has given you through Christ Jesus . | I thank my God always on you , because the grace of God that is given to you through Jesus Christ . | I thank my God have on your people . that , for the grace of God . that is giving you used Jesus Christ . |
| But he that prophesieth speaketh unto men to edification , and exhortation , and comfort . | But a person that prophesies is speaking to people . He gives people strength , encouragement , and comfort . | God , But the person that prophesies is speaking to the people , strengthen stronger and , and comfort . | , then that prophesies speaking to people to make you stronger , and strengthen them , and comfort . |

The phrase-based results are more fluent than the syntax-based results. There are conspicuous changes from archaic words to modern counterparts (e.g., "thee" to "you).

### 5.1.2. Wikipedia

| Wikipedia | Simple Wikipedia | Phrase-based | Syntax-based |
| --- | --- | --- | --- |
| In early 1988 , Balukas gave in to complaints from the men upon her entry to a Chicago based tournament that it was n 't fair she should have the opportunity to play in both divisions when the men only had the opportunity to play | Balukas heard many complaints from the men upon her entry to a Chicago tournament in 1988 . They said it was n 't fair that she got to play on both the men ' s and women ' s side , when the men could only play on | In early 1988 , Balukas gave in to complaints from the men upon her entry to a Chicago based tournament that it was n 't fair she should have the opportunity to play in both divisions when the men only had the chance to play in | In early 1988 , Balukas gave it to discontinue complaints from the men at her entry to a in Chicago , Illinois based tournament that shares it was n &apos;t fair she should have the keeps to play in both into these when the men |
| in one , and withdrew from the men ' s side . | their own side . | one , and withdrew from the men ' s side . | only had the opportunity to play in one , and withdrew from the men &apos; s side . |
| Half points were awarded at the as less than 75 % of the scheduled distance was completed . | Half points were awarded at the Malaysian as less than 75 % of the scheduled distance was completed . | Half points were awarded at the as less than 75 % of the scheduled distance was completed . | Half two points were awarded at the as less than 75 % of the scheduled away , was completed . |
| -LRB- This identifier is not visible to the article ' s reader . -RRB- | -LRB- This identifier is not visible to the article ' s reader . -RRB- | -LRB- This identifier is not visible to the article ' s reader . -RRB- | ( This is a identifier not is visible to the article &apos; his reader . Will Self . ) |
| The North American fauna was a pretty typical boreoeutherian one -LRB- supplemented with Afrotherian proboscids -RRB- . | The North American fauna was typical northern eutheria -LRB- supplemented with Afrotherian proboscids -RRB- . | The North American fauna was a pretty typical boreoeutherian one -LRB- supplemented with Afrotherian proboscids -RRB- . | The Poison Mushroom North \ / their fauna was the bell rang pretty typical boreoeutherian one ( supplemented with it Afrotherian proboscids ) . |

Both the simple text and phrase-based results remain similar to the normal text, but the syntax-based model makes a lot of changes. Again, the phrase-based results are more fluent than the syntax-based results.

## 5.2. MT Components Importance

This experiment aims at examining the importance of each component in the MT models. The phrase-based model has three components—phrase translation, language model, and reordering—and the syntax-based model has two components—phrase translation and language model. I varied the weight of each component and compared the result quality. More specifically, for the phrase-based model, the weights of phrase translation, language model, and reordering were manipulated between {0.2, 1}, {0.5, 1}, and {0.3, 1}, respectively, total 8 combinations. And then for each component, I compared two weights by averaging out the other components. For example, for the weight of phrase translation 0.2, the scores of all combinations where the weight of phrase translation is 0.2 are averaged. For the syntax-based model, the weights of phrase translation and language model were manipulated between {0.2, 0.5, 1} and {0.1, 0.5, 1}, respectively, total 9 combinations. All the scores were calculated on the dev set.

### 5.2.1. Bible

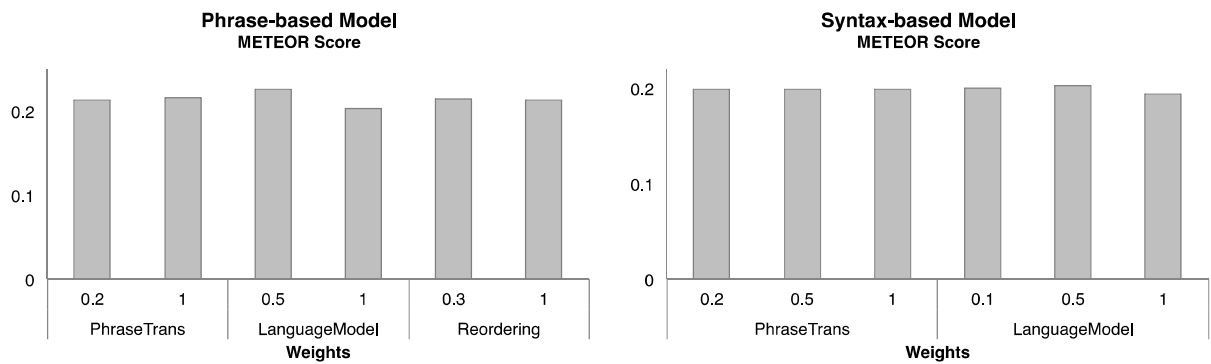

The results show that the weights of phrase translation and reordering have little influence on quality. On the other hand, the weights of language model significantly affect quality. Interestingly, higher weights rather harm results. Since the language models have been trained on the simple training set, the language models may not be general enough.

It may be the case that the difference between 0.2 and 1 is just too small to notice for phrase translation, and the difference between 0.5 and 1 is large enough. Therefore, more weights should be tested to see the overall range of scores. When I tested whether a higher weight of a component yields a higher score for each combination of the other components' weights (i.e., instead of averaging out all combinations), phrase translation constantly showed the trend that higher weights result in higher scores. However, reordering did not; depending on the combination, higher weights yielded both lower scores and higher scores than lower weights. Hence, the hypothesis that reordering would be helpful is not supported. Statistical significance tests may be possible after the scores of more combinations are attained.

### 5.2.2. Wikipedia

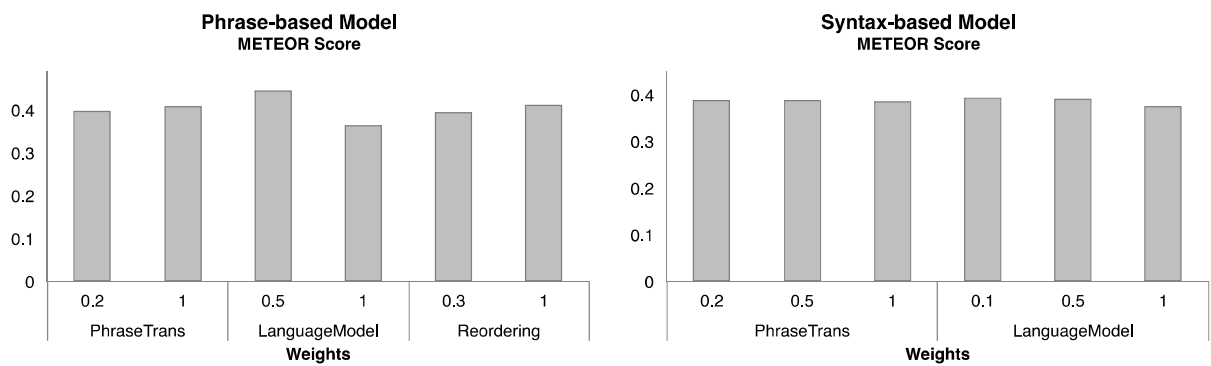

As in the Bible case, phrase translation and reordering have little effect on the result quality. However, language model shows significant influence. Additional analysis shows that for the phrase-based model, higher weights of phrase translation and reordering constantly yield higher scores for each combination of the other weights. Again, the hypothesis that reordering would be harmful is not supported. However, this trend was not found in the syntax-based model. This may indicate that the syntax-based approach is fundamentally defective in Wikipedia. More scores are needed to make more confirmative conclusion.

## 5.3. Performance On Test Set

The weight combination that makes the best score on the dev set was chosen for test on the test set.

### 5.3.1. Bible

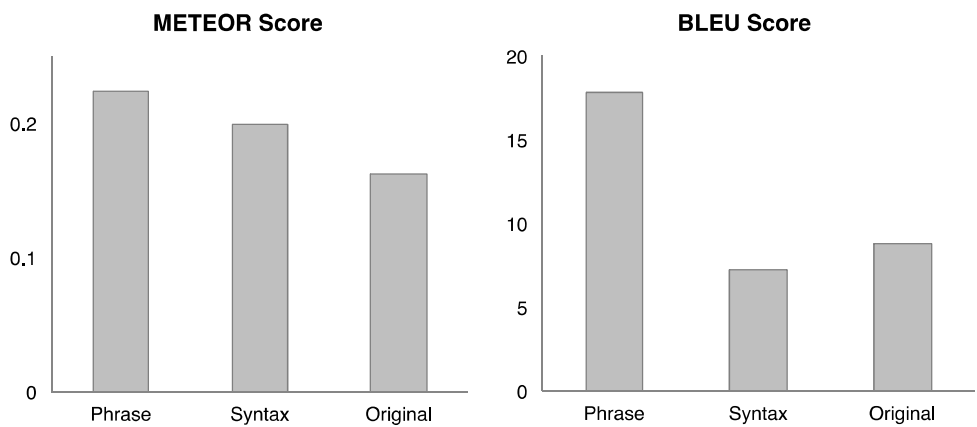

"Original" is the baseline, where the original normal text is tested instead of MT output. That is, this is the score when MT model did nothing and instead output the original text as it is. For METEOR, the phrase-based model performs the best, followed by the syntax-based model and then the original text. BLEU scores show a different result. The phrase-based model still performs the best, but the syntax-based model works worse than the original text. The hypothesis that the syntax-based model would work better than the phrase-based model is not supported.

Next, I looked into the results more in detail using METEOR X-ray. The following graphs show the histogram of sentences in terms of METEOR scores, fragmentation scores, precision, and recall. System-1 is the phrase-based model and System-2 is the syntax-based model.

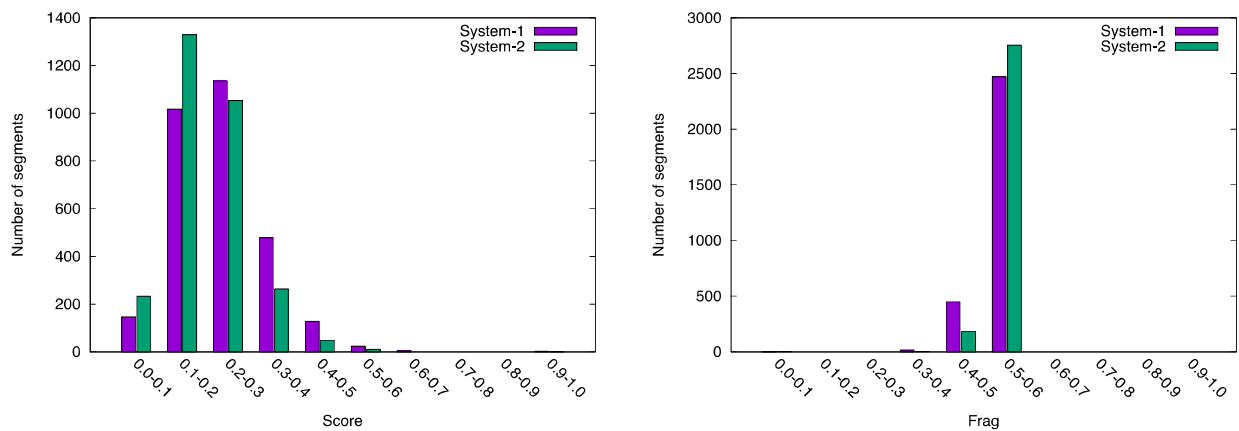

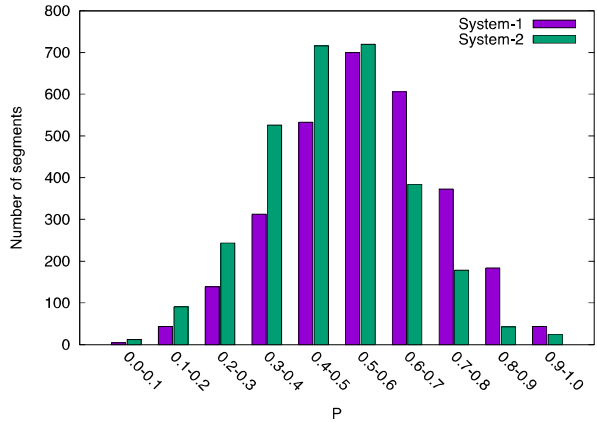 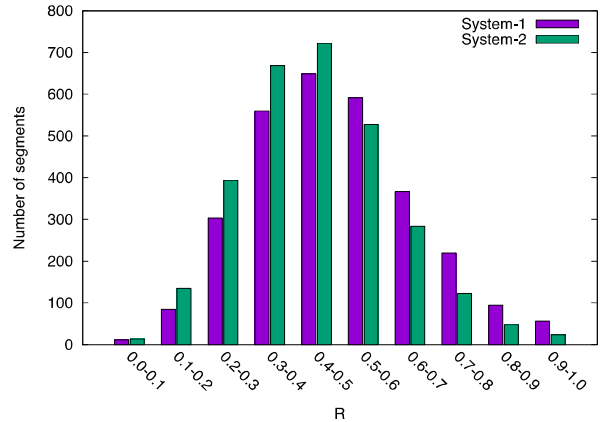

The overall trend is quite straightforward. The phrase-based model tends to have lower scores than the syntax-based model in all metrics. This trend applies to most ranges of sentence length. However, for sentences with 1-10 words, precision shows a quite different pattern as follows.

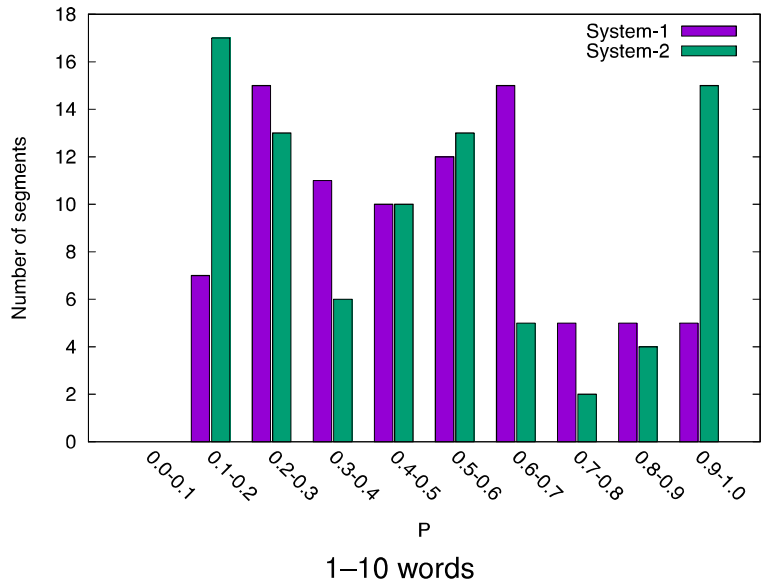

1–10 words

"And the LORD spake unto Moses , saying ,"

| | | the | lord | said | to | moses | , | •○ |
|---|---|---|---|---|---|---|---|---|
| : | | | | | | | | : |
| the | • | | | | | | | the |
| lord | | • | | | | | | lord |
| spoke | | | • | | | | | said |
| to | | | | • | | | | to |
| moses | | | | | • | | | moses |
| . | | | | | | • | | , |
| he | | | | | | | | |
| said | | | | | | | | |
| , | | | | | | | | |

Segment 204

| | | | | | |
|---|---|---|---|---|---|
| P: | 0.625 | vs | 0.909 | : | **0.284** |
| R: | 1.000 | vs | 1.000 | : | **0.000** |
| Frag: | 0.553 | vs | 0.419 | : | **-0.134** |
| Score: | 0.410 | vs | 0.572 | : | **0.162** |

The syntax-based model produces relatively many translations in precision 0.9-1.0. Data analysis shows that this is due to the frequent phrase "And the LORD spake unto Moses , saying ,". This high score barely affects the overall performance, however, because the number of sentences whose length is less than 10 is not many. The example below brings up an important issue about automatic evaluation method. The phrase-based model converted "generations" to "descendants", whereas the syntax-based model converted it to "family history". "Descendants" may be a reasonable choice here, but there is no way to accept it. Also, since the sentence is short, this choice has considerable influence on the result score.

"Now these are the generations of Esau , who is Edom ."

|              | the | family | history | of | esau | . | ●○ |
|--------------|-----|--------|---------|-----|------|---|-----|
| now          |     |        |         |     |      |   | now |
| .            |     |        |         |     |      |   | these |
| these        |     |        |         |     |      |   | men |
| are          |     |        |         |     |      |   | are |
| the          | ●   |        |         |     |      |   | the |
| descendants  |     | ●      |         |     |      |   | family |
| of           |     |        | ●       |     |      |   | history |
| esau         |     |        |         | ●   |      |   | of |
| .            |     |        |         |     | ●    |   | esau |
|              |     |        |         |     |      | ● | . |
|              |     |        |         |     |      |   | he |
|              |     |        |         |     |      |   | is |
|              |     |        |         |     |      |   | edom |
|              |     |        |         |     |      |   | . |

Segment 106

| | | | | | | |
|---|---|---|---|---|---|---|
| P:     | 0.353 | vs | 0.429 | : | **0.076** |
| R:     | 0.500 | vs | 1.000 | : | **0.500** |
| Frag:  | 0.522 | vs | 0.419 | : | **-0.103** |
| Score: | 0.225 | vs | 0.484 | : | **0.259** |

The following graphs compare the length of sentences and edit distance between the normal, simple, phrase-based, and syntax-based text.

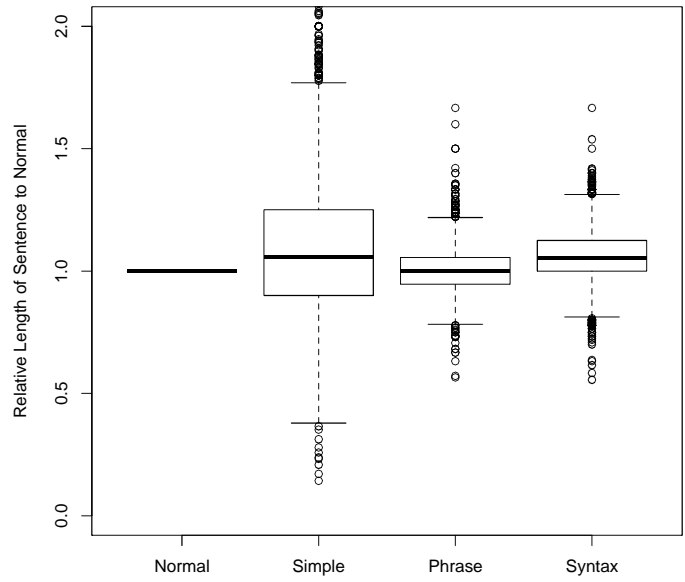

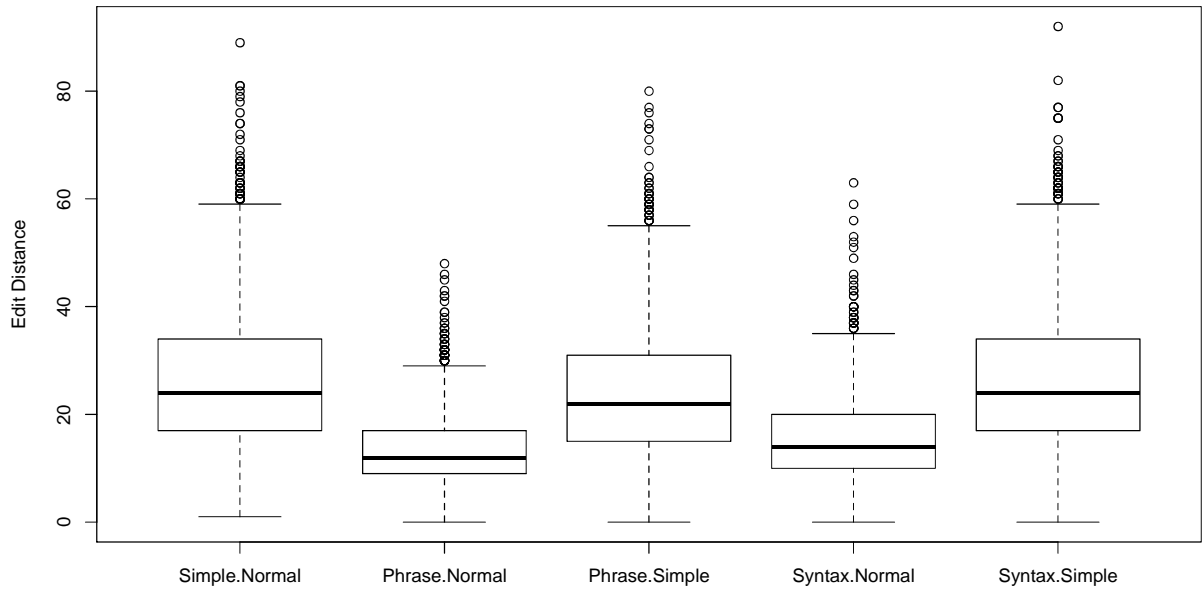

The first box plot shows the relative sentence length of the simple text, phrase-based results, and syntax-based results to the normal text, for each sentence. The simple text is slightly longer than the original text as described before. Both phrase-based model and syntax-based model produce similar length of sentences. The second graph shows that both models are closer to the normal text than to the simple text. This represents the conservative nature of MT models in monolingual translation.

### 5.3.2. Wikipedia

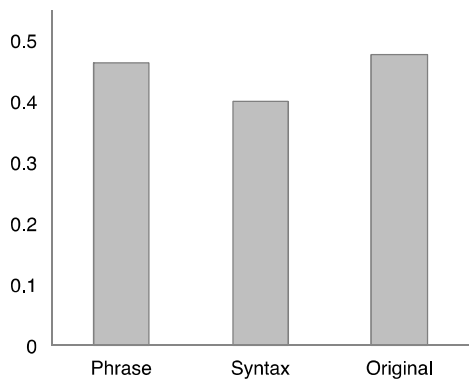
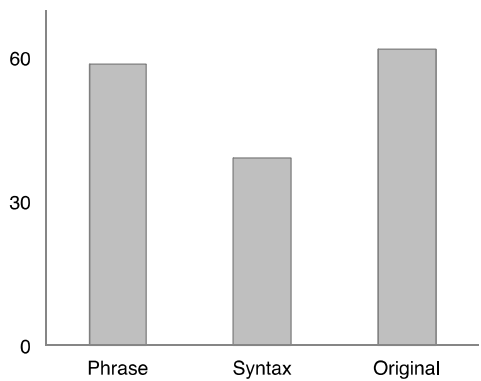

Interestingly, the original text performs the best, followed by the phrase-based model and then the syntax-based model. This is because the original text is too similar to the simplified text. The data statistics in Section 3.2 shows that 30% of the entire sentences remain the same in the simplified version. Generally, scores are higher than the Bible.

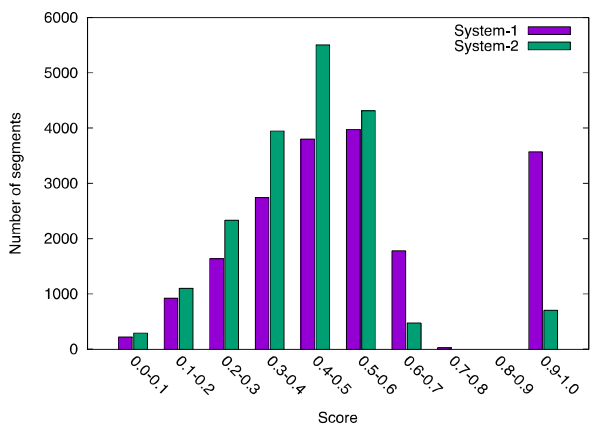
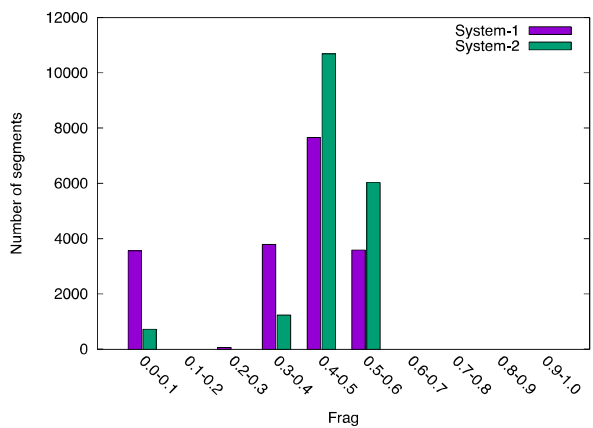
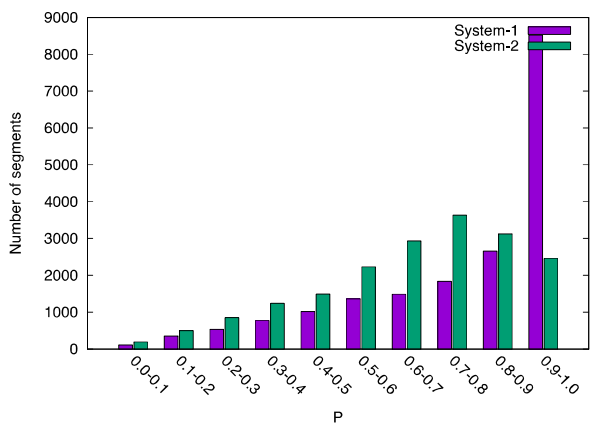
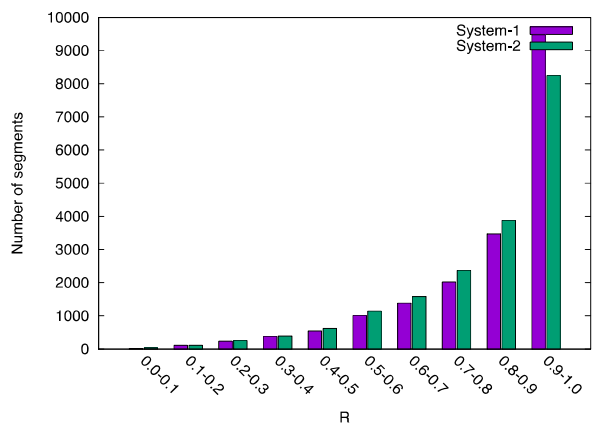

The overall trend of Wikipedia is different from that of the Bible. Scores are skewed toward 1.0 instead of making a bell shape. This is because the simplified text does not require many modifications. This pattern is common over sentences with any length except for those with more than 50 words. The left histogram below shows that both models perform poor on longer sentences, because these sentences require extensive modification. It is confirmed by the right graph, which shows the length of the simple text in the test set in terms of the length of the original text. Longer text gets simplified more intensively.

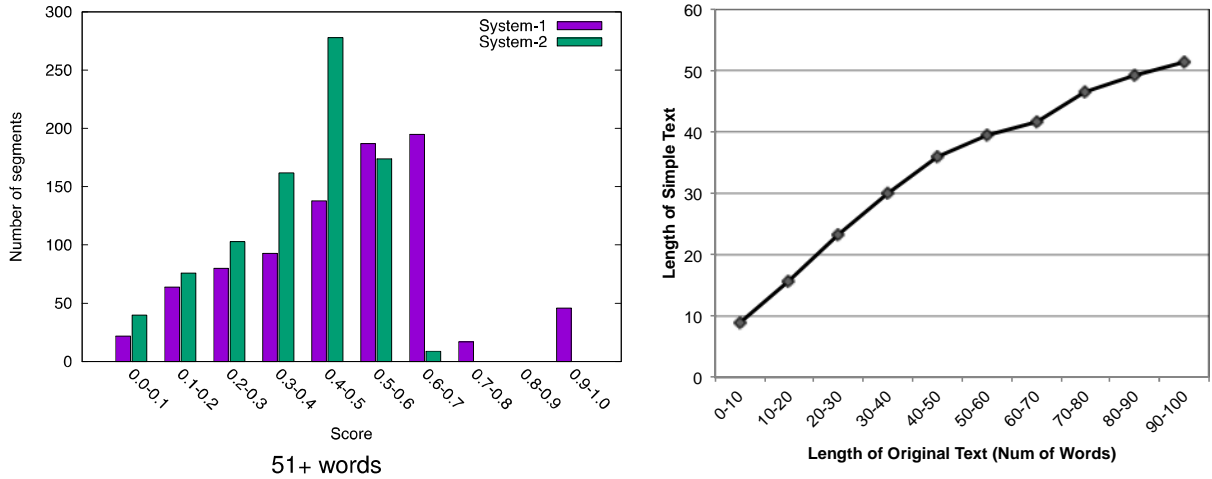

51+ words

One weakness of syntax-based models is their vulnerability to parsing errors. In the example below, "run" is recognized as a noun, which is in turn translated to "the Tennessee Militia".

| | the | exceptions | are | the | 812 | , | mb1 | and | r21 | , | which | run | entirely | within | the | boundary | . | |
|---|---|---|---|---|---|---|---|---|---|---|---|---|---|---|---|---|---|---|
| | | | | | | | | | | | | | | | | | | ●○ |
| the | ● | | | | | | | | | | | | | | | | | the |
| exceptions | | ● | | | | | | | | | | | | | | | | exceptions |
| are | | | ● | | | | | | | | | | | | | | | are |
| the | | | | | | | | | | | | | | | | | | in |
| 812 | | | | ● | | | | | | | | | | | | | | the |
| , | | | | | | | | | | | | | | | | | | muslim |
| mb1 | | | | | | | | | | | | | | | | | | hearts |
| and | | | | | | | ● | | | | | | | | | | | 812 |
| r21 | | | | | | | | ● | | | | | | | | | | , |
| , | | | | | | | | | | ● | | | | | | | | mb1 |
| which | | | | | | | | | | | ● | | | | | | | and |
| run | | | | | | | | | | | | ● | | | | | | r21 |
| entirely | | | | | | | | | | | | | | | | | | margaret |
| within | | | | | | | | | | | | | | | | | | thatcher |
| the | | | | | | | | | | | | | | ● | | | | , |
| boundary | | | | | | | | | | | | | ● | | | | | which |
| . | | | | | | | | | | | | | | | | ● | | the |
| | | | | | | | | | | | | | | | | | | tennessee |
| | | | | | | | | | | | | | | | | | | militia |
| | | | | | | | | | | | | | | | | | | completely |
| | | | | | | | | | | | | | | | | | | inside |
| | | | | | | | | | | | | | | | | | | of |
| | | | | | | | | | | | | | | | | | | the |
| | | | | | | | | | | | | | | | | | | midlands |
| | | | | | | | | | | | | | | | | | ● | . |

Segment 16054

| | | | | | |
|---|---|---|---|---|---|
| P: | 1.000 | vs | 0.412 | : | **-0.588** |
| R: | 1.000 | vs | 0.636 | : | **-0.364** |
| Frag: | 0.000 | vs | 0.514 | : | **0.514** |
| Score: | 1.000 | vs | 0.286 | : | **-0.714** |

Since syntactic translation in the high-level node often produces inaccurate and wordy translations, it would be worth trying to train a tree-to-tree model by allowing only for non-terminal translation.

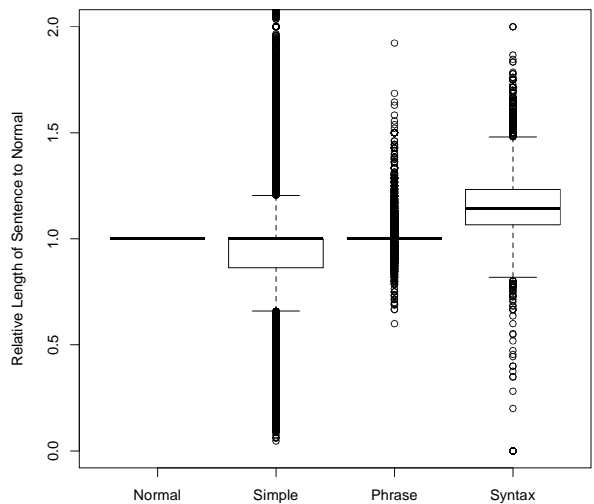

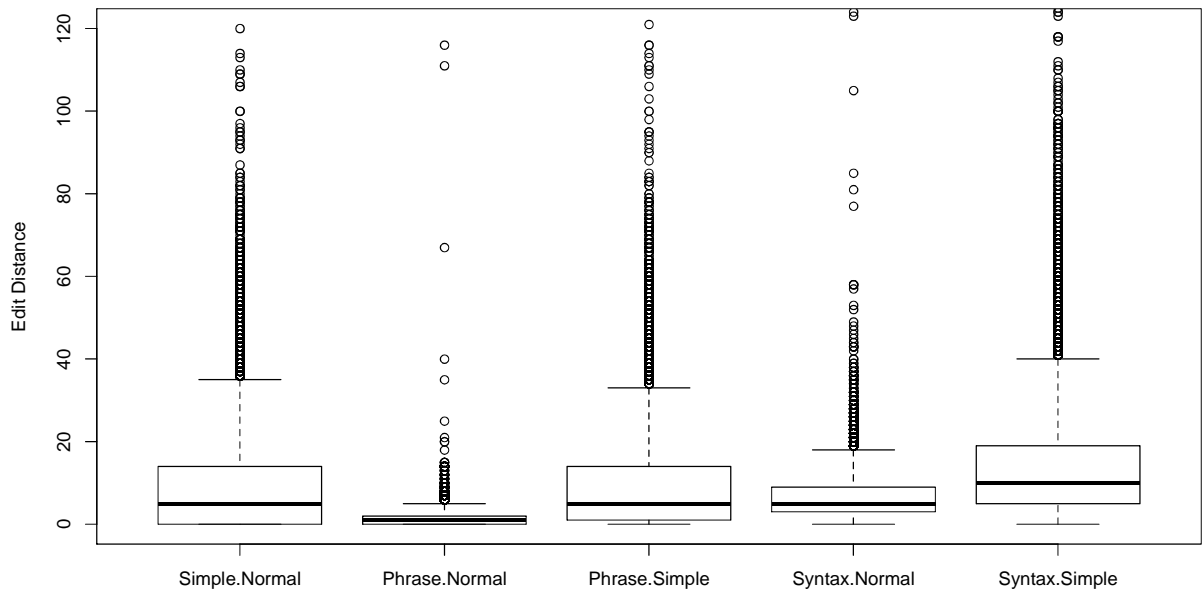

The box plots show interesting characters of the data set. The median length of the simple text is almost the same as the normal text, but the third quartile (50-75%) of the sentences has the same length. This indicates that the simple text is very similar to the original text. The phrase-based model outputs also have almost the same length as the normal sentences, and as shown in the edit distance graph, the phrase-based model does not modify the normal text much. In contrast, the syntax-based model produces wordy sentences, and the variance in length is larger than that of the phrase-based model. Again, both models are closer to the normal text than to the simple text.

5.4. Converting Plain Text to King James Style

This section illustrates an interesting application of MT-based text simplification. So far, text simplification has been accomplished by treating normal text as a source language. However, by training in reverse, we may be complexify easy text as well. Here I show some translation results produced by training on the ERV and the KJV as source and target languages, respectively. This allows us to convert plain text to one with an archaic style. The plain text was obtained from the course website and Aesop's Fables compiled by Project Gutenberg[7].

| Plain Text | Converted King James Style |
|---|---|
| Welcome to Machine Translation ( 11-731 ) . | Blessed to Machine Translation ( 11-731 . ) |

---

[7] http://www.gutenberg.org/

| | |
|---|---|
| This semester we will be using Piazza for class discussion . | this semester we shall be by Piazza for class discussion . |
| The system is designed to get you help fast and efficiently from classmates , Austin , Alon , and me . | the Levitical is designed to thee , and from the swift efficiently classmates , Austin , Alon , and me . |
| Rather than emailing questions to the instructors , we encourage you to post your questions on Piazza . | Rather than emailing questions to the instructors , we comfort ye your questions Piazza on the gallows . |
| Rising up angrily , he caught him and was about to kill him , when the Mouse piteously entreated , saying : If you would only spare my life , I would be sure to repay your kindness . | Rising up , and clamour , and caught him , and he was about to kill him , when the Mouse piteously entreated , saying , If thou wilt save bloodguiltiness my life , that I should repay thy mercy . |
| You ridiculed the idea of my ever being able to help you , not expecting to receive from me any repayment of your favor ; now you know that it is possible for even a Mouse to confer benefits on a Lion . | thou hast ridiculed the pleased of my ever is able to thee , and not aware to receive of me no repayment of thy respect , ye know that it is impossible for a Mouse to confer benefits upon a Lion . |

Some sentences lose the original meaning, but we can see stylistic changes made automatically. I believe the results show the possibility of a general "text synthesizer," in a similar sense to a sound synthesizer or a voice synthesizer. This is not only fun, but also has practical usefulness. For instance, conversion from Standard English to a dialect can be leveraged for educational purposes (Finkelstein et al., 2013). It may also help write special types of documents (e.g., patent) that require certain writing styles.

6. CONCLUSION

This work performed a qualitative analysis of MT-based text simplification on the Bible and Wikipedia. Overall, the Bible produced a poorer quality of translations than Wikipedia, because simplifying the Bible requires significant restructuring of phrases and syntax. On the other hand, Simple English Wikipedia is very similar to the standard Wikipedia already. This also resulted in a poorer quality of translations than doing nothing. There were surprising results that are against initial intuition. First, the sentences produced by the syntax-based model were less fluent (more ungrammatical) than those by the phrase-based model. Parsing errors contribute much to this phenomenon. Also, the syntax-based model performed worse than the phrase-based model for both corpora, which is against my initial expectation that the syntax-based model would perform better for the Bible. The analysis of MT components revealed that emphasizing the language model harms the quality of results. According to my parameter choices, the phrase translation component shows a little improvement in simplification quality. The reordering component, however, shows either no or negative effect for the Bible. This is counter-intuitive because there was apparently considerable difference in phrase orders between the KJV and the ERV.

Data analysis shows that the Bible and Wikipedia have very different nature. This seems to indicate that it is not easy to train a general-purpose text simplification model, because individual domains have unique characteristics, and domain-specific consideration is required.

I planned on investigating new resources of data from the very beginning of this project, but I originally wanted to use novels instead of the Bible. I thought there would be many popular novels (e.g., Shakespeare's) that are translated for adults and edited for kids. This translation is targeting specific grades and carefully examined by educational experts. I expected that these novels might be able to help overcome the lack of parallel corpora for text simplification. However, unfortunately, I could not obtain such novels neither in a digital form nor in a hard copy form. Yet, I think this is still an interesting direction for future study. More consultation with teachers may help obtain those data.

7. REFERENCES

- Coster, W., & Kauchak, D. (2011a). Learning to Simplify Sentences Using Wikipedia. In Proceedings of the Workshop on Monolingual Text-To-Text Generation (pp. 1–9). Portland, Oregon: Association for Computational Linguistics.
- Coster, W., & Kauchak, D. (2011b). Simple English Wikipedia: A New Text Simplification Task. In Proceedings of the 49th Annual Meeting of the Association for Computational Linguistics: Human Language Technologies: Short Papers - Volume 2 (pp. 665–669). Stroudsburg, PA, USA: Association for Computational Linguistics.
- De Belder, J., & Moens, M.-F. (2010). Text simplification for children. SIGIR Workshop on Accessible Search Systems.
- Denkowski, M., & Lavie, A. (2014). Meteor Universal: Language Specific Translation Evaluation for Any Target Language. In Proceedings of the Ninth Workshop on Statistical Machine Translation (pp. 376–380).
- Devlin, S. and Tait, J. (1998). The use of a psycholinguistic database in the simplification of text for aphasic readers. Linguistic Databases (pp. 161–173).
- Dyer, C., Chahuneau, V., & Smith, N. A. (2013). A Simple, Fast, and Effective Reparameterization of IBM Model 2. In Proceedings of the 2013 Conference of the North American Chapter of the Association for Computational Linguistics: Human Language Technologies (pp. 644–648). Atlanta, Georgia: Association for Computational Linguistics.
- Finkelstein, S., Yarzebinski, E., Vaughn, C., Ogan, A., & Cassell, J. (2013) The effects of culturally-congruent educational technologies on student achievement. in Proceedings of Artificial Intelligence in Education (AIED), July 09-13 2013, Memphis, TN.
- Goff, P., II, Farnsley, A. E., & Thuesen, P. J. (2014). The Bible in American Life. Retrieved from http://www.raac.iupui.edu/files/2713/9413/8354/Bible_in_American_Life_Report_March_6_2014.pdf
- Inui, K., Fujita, A., Takahashi, T., Iida, R., and Iwakura, T. (2003). Text simplification for reading assistance: A project note. Proceedings of the Second International Workshop on Paraphrasing. Sapporo, Japan: Association for Computational Linguistics, July 2003, (pp. 9–16).
- Kandula, S., Curtis, D. and Zeng-Treitler, Q. (2010). A semantic and syntactic text simplification tool for health content. AMIA Annual Symposium Proceedings. American Medical Informatics Association (pp. 366–370).

- Kauchak, D. (2013). Improving Text Simplification Language Modeling Using Unsimplified Text Data. In Proceedings of the 51st Annual Meeting of the Association for Computational Linguistics (Volume 1: Long Papers) (pp. 1537–1546). Sofia, Bulgaria: Association for Computational Linguistics.
- Klaper, D., Ebling, S., & Volk, M. (2013). Building a German/Simple German Parallel Corpus for Automatic Text Simplification. In Proceedings of the Second Workshop on Predicting and Improving Text Readability for Target Reader Populations (pp. 11–19). Sofia, Bulgaria: Association for Computational Linguistics.
- Koehn, P., Hoang, H., Birch, A., Callison-Burch, C., Federico, M., Bertoldi, N., … Herbst, E. (2007). Moses: Open Source Toolkit for Statistical Machine Translation. In Proceedings of the 45th Annual Meeting of the Association for Computational Linguistics Companion Volume Proceedings of the Demo and Poster Sessions (pp. 177–180). Prague, Czech Republic: Association for Computational Linguistics.
- Napoles, C., Van Durme, B., and Callison-Burch, C. (2011). Evaluating sentence compression: Pitfalls and suggested remedies. Proceedings of the Workshop on Monolingual Text-To-Text Generation. Portland, Oregon: Association for Computational Linguistics, June 2011 (pp. 91–97).
- Papineni, K., Roukos, S., Ward, T., and Zhu, W. (2002). Bleu: a method for automatic evaluation of machine translation. Proceedings of the 40th Annual Meeting on Association for Computational Linguistics. Stroudsburg, PA, USA: Association for Computational Linguistics (pp. 311–318).
- Petersen, S. E., & Ostendorf, M. (2007). Text simplification for language learners: a corpus analysis. In In Proc. of Workshop on Speech and Language Technology for Education.
- Shardlow, M. (2014). A Survey of Automated Text Simplification. International Journal of Advanced Computer Science and Applications(IJACSA), Special Issue on Natural Language Processing 2014, 58–70. Retrieved from http://dx.doi.org/10.14569/SpecialIssue.2014.040109#sthash.63SgUg00.dpuf
- Stymne, S., Tiedemann, J., Hardmeier, C., & Nivre, J. (2013). Statistical machine translation with readability constraints. In Proceedings of the 19th Nordic Conference on Computational Linguistics (NODALIDA'13) (pp. 375–386).
- Wubben, S., van den Bosch, A., & Krahmer, E. (2012). Sentence Simplification by Monolingual Machine Translation. In Proceedings of the 50th Annual Meeting of the Association for Computational Linguistics (Volume 1: Long Papers) (pp. 1015–1024). Jeju Island, Korea: Association for Computational Linguistics.
- Zhu, Z., Bernhard, D., & Gurevych, I. (2010). A Monolingual Tree-based Translation Model for Sentence Simplification. In Proceedings of the 23rd International Conference on Computational Linguistics (Coling 2010) (pp. 1353–1361). Beijing, China: Coling 2010 Organizing Committee.